\documentclass[10pt,twocolumn,letterpaper]{article} 

\usepackage{avss}
\usepackage{times}
\usepackage{epsfig}
\usepackage{graphicx}
\usepackage{amsmath}
\usepackage{amssymb}

\usepackage{color}

\usepackage{times}  
\usepackage{helvet}  
\usepackage{courier}
\usepackage[hyphens]{url}

\usepackage{caption}

\usepackage{algorithm}
\usepackage{algorithmic}

\usepackage{newfloat}
\usepackage{listings}

\usepackage[table]{xcolor}
\usepackage{amsmath}
\usepackage{amssymb}
\usepackage{booktabs}
\usepackage{array}
\usepackage{soul}
\usepackage{multirow}
\usepackage{subcaption}
\usepackage{xcolor}

\usepackage{pifont}

\renewcommand{\paragraph}[1]{
     \noindent{\textbf{#1}} 
 }

\usepackage{kotex}

\usepackage[capitalize]{cleveref}
\crefname{section}{Sec.}{Secs.}
\Crefname{section}{Section}{Sections}
\Crefname{table}{Table}{Tables}
\crefname{table}{Tab.}{Tabs.}

\usepackage{lipsum}

\avssfinalcopy 

\def\avssPaperID{31} 

\ifavssfinal\fi
\begin{document}

\title{Enhancing Weakly Supervised Video Grounding via Diverse Inference Strategies for Boundary and Prediction Selection}

\author{
Sunoh Kim\\
Computer Engineering\\
Dankook University\\
{\tt\small suno8386@dankook.ac.kr}
\and
Daeho Um\thanks{Corresponding author.}\\
AI Center\\
Samsung Electronics\\
{\tt\small daeho.um@samsung.com}
}

\maketitle

\begin{abstract}
Weakly supervised video grounding aims to localize temporal boundaries relevant to a given query without explicit ground-truth temporal boundaries. While existing methods primarily use Gaussian-based proposals, they overlook the importance of (1) boundary prediction and (2) top-1 prediction selection during inference. In their boundary prediction, boundaries are simply set at half a standard deviation away from a Gaussian mean on both sides, which may not accurately capture the optimal boundaries. In the top-1 prediction process, these existing methods rely heavily on intersections with other proposals, without considering the varying quality of each proposal. To address these issues, we explore various inference strategies by introducing (1) novel boundary prediction methods to capture diverse boundaries from multiple Gaussians and (2) new selection methods that take proposal quality into account. Extensive experiments on the ActivityNet Captions and Charades-STA datasets validate the effectiveness of our inference strategies, demonstrating performance improvements without requiring additional training.


\end{abstract}

\section{Introduction}
Weakly supervised video grounding (WSVG) is a fundamental task in video understanding, aiming to localize relevant temporal boundaries based on a given natural language query, without requiring explicit ground-truth annotations of temporal boundaries~\cite{tan2021logan, wang2021weakly, zheng2022cnm, kim2024gaussian}. This problem has attracted significant attention due to its potential applications in video retrieval~\cite{dong2019dual}, video summarization~\cite{ma2002user}, and human-computer interaction~\cite{tapaswi2016movieqa}. However, the absence of ground-truth annotations presents unique challenges in generating effective temporal boundaries.

\begin{figure*}[t!]
    \centering
    \includegraphics[width=0.9\textwidth]{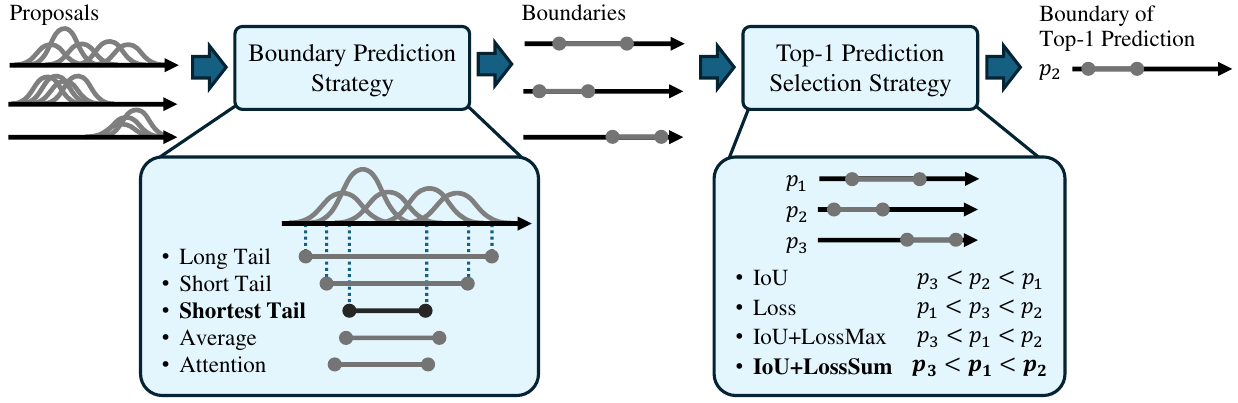}
    \caption{Overview of the proposed inference process, consisting of boundary prediction strategies and top-1 prediction selection strategies. To improve the performance of Gaussian proposal-based methods, we explore five strategies for boundary prediction and four strategies for top-1 prediction selection to derive an optimal combination of inference strategies.}
    \label{fig:pipeline}
\end{figure*}

Previous methods for WSVG have primarily relied on Gaussian-based proposals, where a Gaussian function is used to model temporal moments within a video. 
In early methods, single Gaussian-based proposals~\cite{kong2023dynamic, lv2023counterfactual, yoon2023scanet, zheng2022cnm, zheng2022cpl} provide a structured way to represent temporal moments.
However, the unimodal nature of the single Gaussian proposals limits their ability to capture diverse event occurrences in complex videos. 
To overcome this limitation, Gaussian mixture proposals~\cite{kim2024gaussian} are introduced, generating multiple Gaussians with different weights.
Gaussian mixture proposals enable a comprehensive representation by modeling multiple events, thereby improving the likelihood of capturing relevant temporal moments.

While significant progress has been made in improving proposal generation, the effectiveness of these proposals is ultimately dependent on inference strategies used to determine temporal boundaries and select final predictions.
However, the inference strategies have been largely overlooked in existing methods.
To determine temporal boundaries, most existing methods define the boundaries as half a standard deviation away from the mean of a single Gaussian proposal.
While this approach is somewhat effective, it does not sufficiently explore the diverse range of predictions derived from the multiple Gaussians, potentially leading to suboptimal boundary predictions.
Similarly, the top-1 prediction selection process presents another key challenge in existing methods. Most existing methods rely heavily on intersection-over-union (IoU) scores between proposals to rank and select the best proposal, without considering the semantic relevance of proposals to the given query. As a result, high-IoU proposals may still fail to capture the most meaningful moments in the video.

To address these issues, we explore various inference strategies by proposing two major enhancements:
For boundary prediction, we introduce novel strategies to determine temporal boundaries by leveraging properties of Gaussian mixtures, enabling more precise localization.
For top-1 prediction selection, we introduce novel strategies that assess both IoU and proposal quality, ensuring that the most meaningful proposal is reflected.
Incorporating previous inference strategies from \cite{kim2024gaussian,zheng2022cpl}, we experimentally explore and compare five strategies for boundary prediction and four strategies for top-1 prediction selection, analyzing their effectiveness to derive an optimal combination of inference strategies.

Our key contributions can be summarized as follows:
1) We investigate the underexplored role of inference strategies and identify their limitations in existing Gaussian-based methods.
2) We propose novel boundary prediction and top-1 prediction selection strategies to enhance the temporal localization of Gaussian mixture proposals and capture the most meaningful proposal.
3) Through extensive experiments on the ActivityNet Captions~\cite{krishna2017dense} and Charades-STA~\cite{gao2017tall} datasets, we demonstrate that our inference strategies enhance performance without requiring additional training.

\section{Related Work}
Most weakly supervised video grounding methods can be broadly categorized into two main approaches: sliding window-based methods and reconstruction-based methods. Each approach has distinct strategies for generating temporal proposals.

\paragraph{Sliding window-based methods}~\cite{huang2021cross, tan2021logan, wang2021weakly} rely on a predefined sliding window strategy to generate candidate temporal segments, where the most relevant proposal is selected based on its probability of containing the target moment. 
To enhance the effectiveness of these proposals, \cite{tan2021logan} introduces a multi-level co-attention mechanism that facilitates the learning of visual-semantic representations by integrating contextual dependencies.
Similarly, \cite{huang2021cross} exploits the relationships between multiple sentence queries to capture cross-moment dependencies within videos, thereby improving localization accuracy.
Despite their potential, sliding window-based methods suffer from inefficiencies due to their exhaustive proposal generation.
They produce a large number of overlapping proposals with fixed lengths, necessitating the application of Non-Maximum Suppression (NMS)~\cite{neubeck2006efficient} to eliminate redundancy.
However, this process not only demands substantial computational resources but also relies on prior knowledge about the distribution of temporal segment lengths within each dataset.

\paragraph{Reconstruction-based methods}~\cite{lin2020weakly, zheng2022cnm, zheng2022cpl} approach the problem from a different perspective, assuming that high-quality temporal proposals should be capable of reconstructing the original query from a randomly masked query.
Early work in this category~\cite{lin2020weakly} focuses on capturing and aggregating contextual information from video-sentence pairs to assign scores to proposals sampled across different temporal scales.
However, these methods often face a significant computational burden, as they must select the optimal proposal from a large candidate set, which can be inefficient.
To address this issue, Zheng~\etal~\cite{zheng2022cnm} introduce a learnable Gaussian-based proposal mechanism that reduces the number of candidate proposals while maintaining high localization performance.
This Gaussian proposal framework has since been widely adopted and further extended in multiple studies, including~\cite{kim2024gaussian, kim2024learnable, kong2023dynamic, lv2023counterfactual, yoon2023scanet, zheng2022cpl}.
Kim~\etal~\cite{kim2024gaussian} propose Gaussian mixture proposals, which provide more expressive modeling capabilities compared to the single Gaussian proposals.
However, existing methods largely overlook the role of inference strategies in determining temporal boundaries and selecting the final prediction. In contrast, our work systematically explores and proposes novel inference strategies to enhance the effectiveness of Gaussian-based proposals.

\section{Proposed Method}

For weakly supervised video grounding, we predict temporal boundaries relevant to a given query.
The query-relevant boundaries are defined by the start time $s\in\mathbb{R}$ and end time $e\in\mathbb{R}$.
Adopting Gaussian mixture proposals introduced in \cite{kim2024gaussian}, we generate $N$ Gaussian mixture proposals $\{ \mathbf{P}^{(n)}\}_{n=1}^N$ to predict the query-relevant boundaries.
Let $\mathbf{P}^{(n)}$ denote the $n$-th Gaussian mixture proposal that is represented by a set of Gaussian masks $\mathbf{G}^{(n)}\in \mathbb{R}^{M\times T}$ and corresponding attention weights $\mathbf{a}^{(n)}\in \mathbb{R}^{M}$, which can be written as
\begin{equation}
  \mathbf{P}^{(n)} = \mathbf{G}^{(n)\top}\mathbf{a}^{(n)}\in \mathbb{R}^{T} \text{,}
  \label{eq:gaussian-mixture-proposal}
\end{equation}
where $M$ is the number of Gaussian masks and $T$ is the video length.
Each Gaussian mask is parameterized by its center and width. Specifically, we define the centers and widths of a set of  Gaussian masks $\mathbf{G}$ as vectors: $\mathbf{c}^{(n)} \in \mathbb{R}^{M}$ and $\mathbf{w}^{(n)} \in \mathbb{R}^{M}$, respectively.

To predict temporal boundaries from Gaussian mixture proposals, we employ two inference strategies.
First, we propose boundary prediction strategies to enhance the temporal localization of Gaussian mixture proposals.
Second, we develop top-1 prediction selection strategies to identify the most relevant proposal.
The overall inference process is depicted in \cref{fig:pipeline}.
To determine the optimal combination of inference strategies, we explore five boundary prediction strategies and four top-1 prediction selection strategies.

\subsection{Boundary Prediction}
\label{sec:Boundary Prediction}

In this section, we describe the process of predicting boundaries from generated Gaussian mixture proposals.
First, we compute the left point $\mathbf{l}^{(n)}$ and the right point $\mathbf{r}^{(n)}$ of the $n$-th proposal $\mathbf{P}^{(n)}$ as
\begin{align}
  \mathbf{l}^{(n)}=\mathbf{c}^{(n)}-\frac{\mathbf{w}^{(n)}}{2}\in \mathbb{R}^{M} \text{,} \quad \mathbf{r}^{(n)}=\mathbf{c}^{(n)}+\frac{\mathbf{w}^{(n)}}{2}\in \mathbb{R}^{M} \text{.}
  \label{supp-eq:end}
\end{align}
For computational convenience, we rearrange the elements of \(\mathbf{l}^{(n)}\) and \(\mathbf{r}^{(n)}\) in ascending order:
\begin{math}
  \mathbf{l}_1^{(n)} < \mathbf{l}_2^{(n)} < \dots < \mathbf{l}_M^{(n)}
\end{math}
and 
\begin{math}
  \mathbf{r}_1^{(n)} < \mathbf{r}_2^{(n)} < \dots < \mathbf{r}_M^{(n)} \text{,}
\end{math}
where $\mathbf{l}_m^{(n)}$ and $\mathbf{r}_m^{(n)}$ denote the $m$-th elements of $\mathbf{l}^{(n)}$ and $\mathbf{r}^{(n)}$, respectively.
Then, we derive normalized temporal boundaries, defined by start time $s$ and end time $e$, using the following five strategies: 1) Long Tail, 2) Short Tail, 3) Shortest Tail, 4) Average, and 5) Attention.

\paragraph{Long Tail} strategy predicts temporal boundaries by selecting the left point of the leftmost Gaussian mask and the right point of the rightmost Gaussian mask, which is given by
\begin{align}
  s = \mathrm{max} \left(\mathbf{l}_1^{(n)},0\right) \in\mathbb{R}\text{,} \quad
  e = \mathrm{min} \left(\mathbf{r}_M^{(n)},1\right) \in\mathbb{R}\text{.}
  \label{eq:long-tail}
\end{align}

\paragraph{Short Tail} strategy predicts temporal boundaries by selecting the left point of the second leftmost Gaussian mask and the right point of the second rightmost Gaussian mask, which is given by
\begin{align}
  s = \mathrm{max} \left(\mathbf{l}_2^{(n)},0\right) \in\mathbb{R}\text{,} \quad
  e = \mathrm{min} \left(\mathbf{r}_{M-1}^{(n)},1\right) \in\mathbb{R}\text{.}
  \label{eq:short-tail}
\end{align}

\paragraph{Shortest Tail} strategy predicts temporal boundaries by selecting the left point and right point of the central Gaussian mask, which is given by
\begin{align}
  s = \mathrm{max} \left(\mathbf{l}_{\lfloor \frac{M+1}{2} \rfloor}^{(n)},0\right) \in\mathbb{R}\text{,} \quad
  e = \mathrm{min} \left(\mathbf{r}_{\lfloor \frac{M+1}{2} \rfloor}^{(n)},1\right) \in\mathbb{R}\text{,}
  \label{eq:shortest-tail}
\end{align}
where $\lfloor x \rfloor$ denotes the floor function, which returns the greatest integer less than or equal to $x$.

\paragraph{Average} strategy predicts temporal boundaries by averaging the left points and the right points of all Gaussian masks, respectively, which is given by
\begin{align}
  &s = \mathrm{max} \biggl(\frac{1}{M}\sum_{m=1}^{M}\mathbf{l}_{m}^{(n)},0\biggl) \in\mathbb{R}\text{,}  \\
  &e = \mathrm{min} \biggl(\frac{1}{M}\sum_{m=1}^{M}\mathbf{r}_{m}^{(n)},1\biggl) \in\mathbb{R}\text{.}
  \label{eq:average}
\end{align}

\paragraph{Attention} strategy, which is proposed in \cite{kim2024gaussian}, predicts temporal boundaries by pooling the left points and the right points with attention weights $\mathbf{a}^{(n)}\in \mathbb{R}^{M}$, respectively, which is given by
\begin{align}
 &s = \mathrm{max} (\mathbf{a}^{(n)\top}\mathbf{l}^{(n)},0) \in \mathbb{R} \text{,}  \\
 &e = \mathrm{min} (\mathbf{a}^{(n)\top}\mathbf{r}^{(n)},1) \in \mathbb{R}\text{.} 
  \label{eq:attention}
\end{align}

After applying these boundary prediction strategies, we multiply the normalized temporal boundaries by the video length $T$ to obtain the final predicted temporal boundaries. 

\subsection{Top-1 Prediction Selection}
\label{sec:Top-1 Prediction Selection}

In this section, we detail how to select top-1 prediction from predicted temporal boundaries.
For top-1 prediction selection, we explore four strategies: 1) IoU, 2) Loss, 3) IoU+LossSum, and 4) IoU+LossMax.
Following CPL~\cite{zheng2022cpl}, we design the IoU and Loss strategies.

\paragraph{IoU} strategy, inspired by ensemble learning~\cite{zhou2021ensemble}, leverages the $N$ proposals to vote on the final selection, determining the top-1 prediction.
For each proposal, we compute its IoU with the other $N-1$ proposals and sum these IoU values to obtain its total vote count. The proposal with the highest vote count is then selected as the final prediction.

\paragraph{Loss} strategy utilizes the cross-entropy loss of the reconstructed query from each proposal, which quantifies the semantic relevance of the proposal.
For the $n$-th proposal, the cross-entropy loss $L^{(n)}_{\textrm{ce}}$ of the reconstructed query measures the alignment between the predicted query from the $n$-th proposal and the given query.
Specifically, the reconstruction network~\cite{lin2020weakly} takes the masked query and the proposal-specific video features as input and predicts the query word tokens to reconstruct the original query. 
Then, the cross-entropy loss is computed based on the difference between the predicted and ground-truth query words, ensuring that proposals with higher semantic relevance to the query achieve lower loss values.
Further details on the cross-entropy loss of the reconstructed query can be found in \cite{zheng2022cpl, kim2024gaussian}.
For the final prediction, we select the proposal with the lowest loss.  

\paragraph{IoU+LossSum} strategy combines the IoU and Loss strategies. It calculates the total vote count using the IoU method while incorporating the semantic relevance of each proposal by weighting the $n$-th proposal's IoU with a factor \( w_{\mathrm{sum}}^{(n)} \) based on the cross-entropy loss \( L^{(n)}_{\textrm{ce}} \). The weight is defined as follows:  
\begin{equation}
    w_{\mathrm{sum}}^{(n)} = 1 - \frac{L^{(n)}_{\textrm{ce}}}{\sum_{n=1}^{N}L^{(n)}_{\textrm{ce}}}
\end{equation}

\paragraph{IoU+LossMax} strategy follows a similar approach to IoU+LossSum but uses a different weighting scheme. Instead of normalizing the cross-entropy loss by its sum across all proposals, this strategy normalizes it by the maximum loss value among the proposals. The weight is given by:  
\begin{equation}
    w_{\mathrm{max}}^{(n)} = 1 - \frac{L^{(n)}_{\textrm{ce}}}{\max\limits_{1 \leq n \leq N} L^{(n)}_{\textrm{ce}}
}
\end{equation}  
This weighting scheme ensures that proposals with the highest loss contribute minimally to the vote count, further emphasizing the selection of semantically relevant proposals.

\section{Experiments}
\begin{table*}[t!]
  \centering
  \caption{Performance comparisons of boundary prediction and top-1 prediction selection strategies.} 
  \resizebox{0.85\linewidth}{!}
  {
  \begin{tabular}{llrrrr|rrrr}
\toprule
\multirow{2}{*}{\shortstack[l]{Boundary \\ Prediction}} & \multirow{2}{*}{\shortstack[l]{Top-1 Proposal \\ Selection}} & \multicolumn{4}{c|}{Charades-STA} & \multicolumn{4}{c}{ActivityNet Captions} \\
\cmidrule(lr){3-6} \cmidrule(lr){7-10}
& &  IoU@0.3 &  IoU@0.5 &  IoU@0.7 &  mIoU &  IoU@0.3 &  IoU@0.5 &  IoU@0.7 &  mIoU \\
\midrule
\multirow{4}{*}{Long Tail} & IoU        & 53.01 & 35.81 & 18.52 & 35.22 & 55.70 & 30.18 & 12.32 & 37.02 \\
& Loss & 66.53 & 48.89 & 22.83 & 43.47 & 48.17 & 26.95 & 10.67 & 31.48 \\
& IoU+LossMax & \textbf{69.41} & 51.23 & 25.78 & 45.94 & 53.54 & 29.50 & 10.93 & 35.06 \\
& IoU+LossSum & 57.60 & 40.03 & 20.61 & 38.36 & 55.71 & 30.25 & 12.34 & 37.04 \\
\midrule
\multirow{4}{*}{Short Tail} & IoU       & 52.72 & 36.10 & 18.87 & 35.20 & 55.54 & 30.20 & 12.35 & 37.00 \\
& Loss & 66.50 & 48.96 & 22.93 & 43.48 & 48.11 & 26.99 & 10.65 & 31.47 \\
& IoU+LossMax & 69.35 & 51.42 & 25.84 & 45.91 & 53.55 & 29.53 & 10.99 & 35.07 \\
& IoU+LossSum & 56.78 & 40.03 & 20.61 & 38.01 & 55.65 & 30.27 & 12.37 & 37.03 \\
\midrule
\multirow{4}{*}{Shortest Tail} & IoU        & 52.98 & 35.94 & 18.75 & 35.25 & 55.86 & 30.21 & 12.37 & 37.09 \\
& Loss & 66.47 & 48.92 & 22.86 & 43.47 & 48.16 & 26.96 & 10.69 & 31.48 \\
& IoU+LossMax & 69.32 & 51.39 & 25.90 & \textbf{45.95} & 53.57 & 29.49 & 10.99 & 35.07 \\
& IoU+LossSum & 56.62 & 39.39 & 20.39 & 37.79 & \textbf{55.89} & \textbf{30.30} & \textbf{12.38} & \textbf{37.11} \\
\midrule
\multirow{4}{*}{Average} & IoU        & 52.63 & 35.94 & 19.19 & 35.17 & 55.78 & 30.19 & 12.30 & 37.05 \\
& Loss & 66.37 & 49.02 & 23.05 & 43.47 & 48.06 & 26.91 & 10.64 & 31.43 \\
& IoU+LossMax & 69.25 & \textbf{51.61} & \textbf{25.93} & 45.92 & 53.57 & 29.52 & 10.96 & 35.06 \\
& IoU+LossSum & 56.62 & 39.87 & 21.12 & 37.91 & 55.84 & 30.29 & 12.34 & 37.08 \\
\midrule
\multirow{4}{*}{Attention} & IoU        & 52.66 & 35.97 & 19.22 & 35.19 & 55.79 & 30.20 & 12.30 & 37.05 \\
& Loss & 66.37 & 49.02 & 23.05 & 43.47 & 48.06 & 26.91 & 10.64 & 31.43 \\
& IoU+LossMax & 69.22 & 51.58 & \textbf{25.93} & 45.90 & 53.57 & 29.52 & 10.96 & 35.06 \\
& IoU+LossSum & 56.62 & 39.90 & 21.12 & 37.90 & 55.85 & 30.28 & 12.34 & 37.08 \\
\bottomrule
  \end{tabular}}
  \label{tab:inference_results}
\end{table*}

\begin{table*}[t!]
  \centering
  \caption{Performance comparisons with previous Gaussian proposal-based methods. The best and second-best results are represented as bold and underlined numbers, respectively.} 
  \resizebox{0.75\linewidth}{!}
  {
  \begin{tabular}{l|cccc|cccc}
    \toprule
    \multirow{2}{*}{Method} & \multicolumn{4}{c|}{Charades-STA} & \multicolumn{4}{c}{ActivityNet Captions} \\
    & IoU@0.3 & IoU@0.5 & IoU@0.7 & mIoU & IoU@0.3 & IoU@0.5 & IoU@0.7 & mIoU \\
    \midrule
    CNM~\cite{zheng2022cnm}
    & 60.39 & 35.43 & 15.45 & - & 55.68 & 33.33 & - & - \\
    CPL~\cite{zheng2022cpl} & 66.40 & 49.24 &
    22.39 & 43.48 & 55.73 & \underbar{31.37} & - & - \\
    CPI~\cite{kong2023dynamic} & 67.64 & 50.47 &
    24.38 & - & - & - & - & - \\
    CCR~\cite{lv2023counterfactual} & 68.59 & 50.79 & 23.75 & 44.66 & 53.21 & 30.39 & - & 36.69 \\
    SCANet~\cite{yoon2023scanet} & 68.04 & 50.85 & 24.07 & - &  \underbar{56.07} & \textbf{31.52} & - & - \\
    PPS~\cite{kim2024gaussian} & \underbar{69.06} & \textbf{51.49} &
    \textbf{26.16} & - & \textbf{59.29} & 31.25 & - & \textbf{37.59} \\
    \midrule
    Baseline & 66.37 & 49.02 & 23.05 & 43.47 & 48.06 & 26.91 & 10.64 & 31.43 \\
    Ours & \textbf{69.32} & \underbar{51.39} & \underbar{25.90} & \textbf{45.95} & 55.89 & 30.30 & \textbf{12.38} & \underbar{37.11} \\
    \bottomrule
  \end{tabular}}
  \label{tab:comparisons}
\end{table*}

\begin{figure}[t!]
    \centering
    \includegraphics[width=\linewidth]{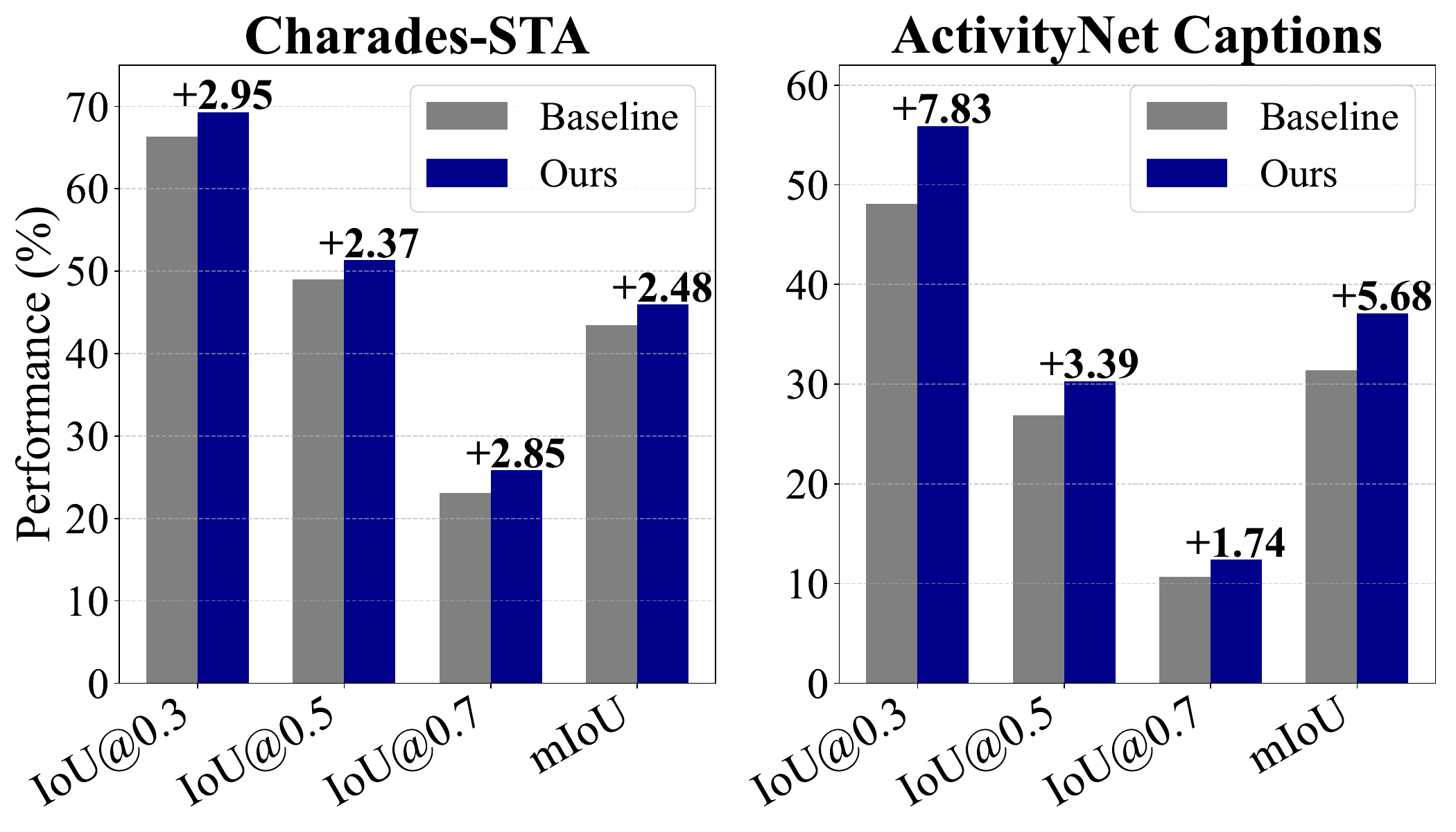}
    \caption{Performance improvements of our method over the baseline.}
    \label{fig:comparisons}
\end{figure}

\subsection{Datasets}
\label{sec:datasets}
\paragraph{ActivityNet Captions.}
The ActivityNet Captions dataset~\cite{krishna2017dense} consists of 37,417, 17,505, and 17,031 video-sentence pairs for training.
Video segment features are extracted using C3D~\cite{tran2015learning}.

\paragraph{Charades-STA.}
The Charades-STA dataset~\cite{gao2017tall} comprises 16,128 video-sentence pairs sourced from 6,672 videos, split into 12,408 pairs for training and 3,720 for testing.
Video segment features are extracted using I3D~\cite{carreira2017quo}.

\subsection{Experimental Setting}
\label{sec:experimental-setting}

\paragraph{Evaluation metrics.} 
We follow the evaluation criteria established in \cite{gao2017tall} and utilize two metrics.
First, IoU=$m$ measures the proportion of cases where at least one of the top-1 predicted temporal boundaries has an Intersection over Union (IoU) with the ground truth exceeding $m$. 
Second, mIoU represents the mean of the highest IoU values among the top-1 predicted temporal boundary.


\paragraph{Implementation details.}
The maximum number of video segments is 200, while the sentence query length is capped at 20.
For transformer-based models, we employ three-layer transformers with four attention heads.
The feature dimensions are fixed at 256.
We follow Kim~\etal~\cite{kim2024gaussian} to generate Gaussian mixture proposals.
Training is conducted using the Adam optimizer~\cite{kingma2014adam}.
The learning rate is set to 0.0004, with a mini-batch size of 32.

\subsection{Ablation Study}
\label{sec:ablation-study}

To evaluate the effectiveness of our proposed inference strategies, we conduct an ablation study by comparing different boundary prediction and top-1 prediction selection strategies. Table~\ref{tab:inference_results} summarizes the experimental results on Charades-STA and ActivityNet Captions datasets.

\paragraph{Effect of Boundary Prediction Strategies.}  
We first analyze the impact of different boundary prediction strategies on the localization performance. Among the five strategies, {IoU+LossMax with Shortest Tail} achieves the highest mIoU of \textbf{45.95\%} on Charades-STA, while {IoU+LossSum with Shortest Tail} performs best on ActivityNet Captions with a mIoU of \textbf{37.11\%}. 
These findings suggest that selecting the central Gaussian mask (Shortest Tail) more effectively identifies query-relevant boundaries than the Long Tail and Short Tail strategies. The reason is that broader temporal boundaries are more likely to include redundant moments, reducing localization precision.

\paragraph{Effect of Top-1 Prediction Selection Strategies.}  
We further examine the role of different top-1 prediction selection strategies. {IoU+LossMax consistently outperforms other strategies across different boundary prediction methods}, demonstrating that integrating both proposal agreement (IoU) and semantic relevance (Loss) is beneficial. Compared to the standard IoU-based voting, IoU+LossMax \textbf{improves IoU@0.5 by up to 15.42\%} (from 35.81\% to 51.23\%) on Charades-STA when combined with the Long Tail strategy. Similarly, it achieves \textbf{the best IoU@0.7 of 25.93\%} when used with the Shortest Tail or Attention strategy. This suggests that mitigating the influence of high-loss proposals leads to better selection of the most query-relevant boundaries.

Meanwhile, {IoU+LossSum also demonstrates strong performance, particularly on ActivityNet Captions}, achieving the highest mIoU of \textbf{37.11\%} when combined with the Shortest Tail strategy. Unlike IoU+LossMax, which suppresses high-loss proposals by normalizing with the maximum loss value, IoU+LossSum normalizes with the total loss sum, providing a more balanced weighting across all proposals. This approach may be beneficial when proposals exhibit relatively uniform loss distributions, as it prevents overly penalizing moderately high-loss proposals that may still contain useful information. The strong performance of IoU+LossSum on ActivityNet Captions suggests that this dataset benefits from a more evenly distributed proposal weighting strategy compared to Charades-STA.

\paragraph{Optimal Combination}  
The experimental results show that the optimal configurations are Shortest Tail for boundary prediction combined with IoU+LossMax for selection on Charades-STA and IoU+LossSum for selection on ActivityNet Captions.
These results imply two key points.
First, selecting the central Gaussian mask helps avoid unnecessary moments and captures the most query-relevant boundaries.
Second, leveraging a loss-aware voting mechanism reflects the semantic relevance of proposals.

\subsection{Comparison with Other Methods}
\label{sec:comparison-with-state-of-the-art-methods}

To validate the effectiveness of our proposed method, we compare it with existing Gaussian proposal-based methods on the Charades-STA and ActivityNet Captions datasets, as shown in \cref{tab:comparisons}.
For the baseline, we use Attention~\cite{kim2024gaussian} for boundary prediction and Loss~\cite{zheng2022cpl} for top-1 prediction selection.
For our method, we employ Shortest Tail for boundary prediction and IoU+LossMax for top-1 prediction selection on Charades-STA, while utilizing Shortest Tail and IoU+LossSum for ActivityNet Captions, respectively.
On the Charades-STA dataset, our method achieves \textbf{69.32\%} at IoU=0.3, \textbf{51.39\%} at IoU=0.5, and \textbf{25.90\%} at IoU=0.7, demonstrating an improvement of \textbf{+2.95\%}, \textbf{+2.37\%}, and \textbf{+2.85\%} over the baseline, respectively, as shown in \cref{fig:comparisons}. Notably, our method outperforms all compared methods at IoU=0.3 and mIoU, as shown in \cref{tab:comparisons}. This result highlights our method’s ability to generate high-quality temporal proposals, especially under less strict IoU thresholds and average IoU.
On the ActivityNet Captions dataset, our method achieves \textbf{55.89\%} at IoU=0.3, \textbf{30.30\%} at IoU=0.5, and \textbf{12.38\%} at IoU=0.7, demonstrating a large improvement of \textbf{+7.83\%}, \textbf{+3.39\%}, and \textbf{+1.74\%} over the baseline, respectively, as shown in \cref{fig:comparisons}.
Our method is competitive with other state-of-the-art methods, as shown in \cref{tab:comparisons}. Moreover, our model achieves \textbf{37.11\%} in mIoU, securing the second-best performance.

\section{Conclusion}
In this work, we investigated the underexplored role of inference strategies in weakly supervised video grounding (WSVG) and proposed novel approaches to enhance the effectiveness of Gaussian-based proposals. While previous methods primarily focused on improving proposal generation, we demonstrated that boundary prediction and top-1 proposal selection play a crucial role in accurately localizing target moments. To this end, we introduced novel boundary prediction strategies and top-1 selection strategies. 
Notably, Shortest Tail strategy effectively determines query-relevant boundaries by selecting the central Gaussian mask, avoiding unnecessary moments.
Moreover, IoU+LossMax and IoU+LossSum integrate proposal agreement with semantic relevance, leading to more accurate top-1 predictions.
Through extensive experiments, we showed that our inference strategies significantly improve grounding accuracy without requiring additional training.
Our findings present a practical and efficient solution for real-world video surveillance applications, improving localization accuracy without any additional training overhead. 
This highlights the importance of effective inference strategies, paving the way for future research on WSVG, particularly in real-time and resource-constrained environments.


{\small
\bibliographystyle{ieee}
\bibliography{reference}
}

\end{document}